\pdfoutput=1

\documentclass[11pt]{article}

\usepackage[]{acl}
\usepackage{multirow}
\usepackage{makecell}
\usepackage{times}
\usepackage{latexsym}
\usepackage{amssymb} 
\usepackage[T1]{fontenc}

\usepackage[utf8]{inputenc}

\usepackage{microtype}
\usepackage{CJKutf8}        
\usepackage{xcolor}

\usepackage{inconsolata}
\usepackage{booktabs}
\usepackage{graphicx}

\usepackage{algorithm}
\usepackage{algpseudocode}
\usepackage{amsmath}
\usepackage{xspace}

\definecolor{codegreen}{rgb}{0,0.3,0.6}
\definecolor{codegray}{rgb}{0.5,0.5,0.5}

\newcommand{\ie}{\emph{i.e.,}\xspace}

\newcommand{\paratitle}[1]{\vspace{1.5ex}\noindent\textbf{#1}}

\newcommand{\ignore}[1]{}

\usepackage{tcolorbox}
\definecolor{darkorange}{RGB}{255, 140, 0}

\definecolor{lightgreen}{RGB}{145, 204, 117}
\definecolor{lightyellow}{RGB}{250, 200, 88}
\definecolor{lightred}{RGB}{238, 102, 102}
\definecolor{lightblue}{RGB}{115, 192, 222}
\newtcolorbox{promptbox}[3][Judge Prompt]{
colback=black!5!white,
arc=5pt, 
boxrule=0.5pt,
fonttitle=\bfseries,
title=#1, 
before upper={\small}, fontupper=\fontfamily{ptm}\selectfont,
colframe=#2,
label=#3,
}

\title{R1-Searcher++: Incentivizing the Dynamic Knowledge Acquisition of LLMs via Reinforcement Learning}

\author{
    \textbf{Huatong Song\textsuperscript{{1}}\thanks{\llap{}\:\:\:Equal contributions.},
            Jinhao Jiang\textsuperscript{{1}}\footnotemark[1],
            Wenqing Tian\textsuperscript{{3}},
            Zhipeng Chen\textsuperscript{{1}},
            }\\
    \textbf{Yuhuan Wu\textsuperscript{{1}},
            Jiahao Zhao\textsuperscript{{1}},
            Yingqian Min\textsuperscript{{1}},
            }\\
    \textbf{Wayne Xin Zhao\textsuperscript{{1}}\thanks{\llap{}\:\:\:Corresponding author.},
            Lei Fang\textsuperscript{{2}},
            Ji-Rong Wen\textsuperscript{{1}}}\\
	\textsuperscript{1}Gaoling School of Artificial Intelligence, Renmin University of China.\\
	\textsuperscript{2}DataCanvas Alaya NeW.
    \textsuperscript{3}Beijing Institute of Technology.\\
    \texttt{\{songhuatong123, jiangjinhao\}@ruc.edu.cn, batmanfly@gmail.com}\\
}

\begin{document}
\maketitle
\begin{abstract}
Large Language Models (LLMs) are powerful but prone to hallucinations due to static knowledge. Retrieval-Augmented Generation (RAG) helps by injecting external information, but current methods often are costly, generalize poorly, or ignore the model’s internal knowledge.
In this paper, we introduce R1-Searcher++, a novel framework designed to train LLMs to adaptively leverage both internal and external knowledge sources. R1-Searcher++ employs a two-stage training strategy: an initial SFT Cold-start phase for preliminary format learning, followed by RL for Dynamic Knowledge Acquisition. The RL stage uses outcome-supervision to encourage exploration, incorporates a reward mechanism for internal knowledge utilization, and integrates a memorization mechanism to continuously assimilate retrieved information, thereby enriching the model's internal knowledge. By leveraging internal knowledge and external search engine, the model continuously improves its capabilities, enabling efficient retrieval-augmented reasoning.
Our experiments demonstrate that R1-Searcher++ outperforms previous RAG and reasoning methods and achieves efficient retrieval.
The code is available at \url{https://github.com/RUCAIBox/R1-Searcher-plus}.

\end{abstract}

\section{Introduction}\label{sec:intro}

Large language models (LLMs)~\citep{zhao2023survey} have demonstrated remarkable reasoning capabilities by only leveraging the information encoded in their parameters. However, their reliance on static, internal knowledge leads to notable limitations. At the simultaneously, this reliance easily leads to  hallucinations~\citep{Huang_2025}, so LLMs may struggle with open-ended tasks~\citep{wang2025omnievalomnidirectionalautomaticrag, trivedi2022musique}. Therefore, it is crucial to enable LLMs to access external information when they are confused during the reasoning process to achieve more deliberative reasoning~\citep{rag-star}.

To address this issue, extensive research has focused on augmenting LLMs with external information sources (\ie RAG~\citep{gao2024retrievalaugmentedgenerationlargelanguage}). 
Early approaches emphasize specific prompting strategies to guide LLMs~\citep{li2025searcho1agenticsearchenhancedlarge, teng2025atomthoughtsmarkovllm} and subsequent studies investigate to distill this capability into smaller LLMs through supervised fine-tuning (SFT)~\citep{corag}. However, recent findings suggest that SFT-based distillation can cause models to memorize solution paths, limiting their generalization to novel scenarios~\citep{chu2025sftmemorizesrlgeneralizes}. Further proposals include a test-time scaling method~\citep{li2024can}, notably employing the Monte Carlo Tree Search (MCTS) framework~\citep{sun2025rearterretrievalaugmentedreasoningtrustworthy} to enhance solution-finding by expanding the search space during inference, but this approach incurs significant inference overhead, reducing its practicality for widespread use. 
Recent studies employ end-to-end outcome-based reinforcement learning (RL) to train models, enabling them to autonomously explore external retrieval environments during inference~\citep{search-r1, r1-searcher}. This approach fosters the development of self-directed retrieval capabilities in LLMs as they reason.
However, such models often become overly reliant on external search engines after training, neglecting the utilization of their internal knowledge.

In practice, when humans attempt to solve factual questions, they first recall their internal knowledge, and only turn to search engines when they recognize a lack of information. At the same time, after obtaining the external searched information, humans would memorize this knowledge for future use. For LLMs, extensive pretraining on large-scale data has already endowed them with substantial internal knowledge~\citep{qwen2025qwen25technicalreport}. Therefore, it is essential to equip models with the ability to dynamically switch between internal and external knowledge sources as needed. Furthermore, models should be encouraged to effectively memorize useful information encountered during training~\citep{jiang2024long}, progressively enriching their internal knowledge and continuously evolving toward greater intelligence.

In this paper, we present \textbf{R1-Searcher++}, a novel framework designed to teach LLMs to adaptively leverage both internal and external knowledge. 
We adopt a two-stage training strategy: \emph{SFT Cold-start} and \emph{RL for Dynamic Knowledge Acquisition}. In the first phase, we employ reject sampling to collect data that meets the format requirements and perform a cold start with SFT. In the second stage, we further train the model using outcome-based RL to guide the model in dynamically acquiring knowledge, which is to encourage reliance on internal knowledge when confident, and invoke external search mechanisms when uncertain, based on a carefully designed reward design. Additionally, we further introduce a memory mechanism, enabling the model to retain knowledge encountered during training by converting and memorizing retrieved content. This mechanism continuously enriches its internal knowledge, empowering it to effectively balance internal reasoning and external retrieval through autonomous exploration and timely memorization.

To verify the effectiveness of R1-Searcher++, we conduct extensive experiments based on Qwen-2.5-7B-Instruct. Notably, our method surpasses the strong baseline by up to 4.3\% and reduces the retrieval count by 42.9\% compared to vanilla RL-based approaches.

Our principal contributions are as follows:

$\bullet$ We introduce R1-Searcher++, teaching LLMs to adaptively leverage both internal and external knowledge through a two-stage training strategy.

$\bullet$We encourage the model to actively leverage its internal knowledge while efficiently memorizing external information, enabling dynamic knowledge acquisition through exploration and memorization.

$\bullet$ Extensive experiments show that R1-Searcher++ outperforms existing RAG methods, while significantly reducing the number of retrievals compared to vanilla RL-based approaches.

\section{Related Work}

\paratitle{Retrieval-Augmented Generation.} To improve the factual accuracy of LLM inference and reduce hallucinations, researchers have proposed enhancing language models by incorporating external information sources, a paradigm known as RAG~\citep{ragsurvey}. Early RAG approaches primarily include Branching~\citep{kim2024sure}, Summarization~\citep{li-etal-2023-compressing}, and Adaptive Retrieval~\citep{jeong2024adaptive} strategies.
As foundation models have become increasingly capable, exhibiting strong CoT reasoning abilities, many studies have combined RAG with CoT. These efforts include methods that prompt the model to perform step-by-step retrieval~\citep{shao2023enhancing, trivedi2023interleaving} and strategies that distill such capabilities into smaller LLMs~\citep{self-rag}. In parallel, several works have explored test-time scaling, notably using  MCTS~\citep{feng2025airrag} to dynamically expand reasoning paths. However, such approaches often incur substantial inference-time overhead.
More recently, researchers have trained models using outcome- supervision RL~\citep{zheng2025deepresearcher} to encourage the exploration of more effective actions and retrieval behaviors, but it leads models to over-rely on external search engines, diminishing their ability to leverage internal knowledge~\citep{wang2025otc}.
Enabling LLMs to effectively integrate and alternate between internal knowledge and external retrieval remains a significant challenge.

\paratitle{Reinforcement Learning.}
To improve training efficiency, several off-policy algorithms have since been proposed~\citep{dpo, ethayarajh2024kto}; however, these methods still face limitations in terms of preference modeling accuracy and generalization capability~\citep{pang2024iterative}.
DeepseekMath introduced the GRPO algorithm~\citep{shao2024deepseekmathpushinglimitsmathematical}, which enables efficient self-exploration through a mechanism of relative preference optimization. Building on this, Deepseek-R1~\citep{deepseekr1} have demonstrated that outcome-based RL can significantly enhance the reasoning abilities of large models.
More recently, studies have begun to investigate RL algorithms specifically designed to improve LLM reasoning capabilities~\citep{yu2503dapo, yuan2025vapo}. In parallel, other research efforts have applied RL to the retrieval domain, aiming to enable deep search capabilities~\citep{chen2025research}. However, the use of RL that combine LLM-driven retrieval and reasoning remains largely simplistic and underexplored.




\section{Preliminary}





To enhance the performance of LLMs in open-domain multi-hop question answering tasks~\citep{ho2020constructing}, in this work, we focus on enabling the model to autonomously decide when to use its internal knowledge or to invoke an external retriever to answer the given questions with the LLM self-improving paradigm, which can improve both reasoning effectiveness and efficiency.
To this end, we introduce three special tokens to format the LLM reasoning process, \ie \texttt{<internal>}, \texttt{<external>}, and \texttt{<document>}. 
Concretely, during the reasoning process, the LLM with parameters $\theta$ determines whether the current step requires external knowledge to help perform reasoning.
If so, it triggers the \texttt{<external>} to issue a $query_t$, which is sent to a retriever to retrieve the top-$K$ relevant documents $D_t=\{d_{t,k}\}^{K}_{k=1}$ from an external corpus. 
These retrieved documents are incorporated into the reasoning path with another special token \texttt{<document>}.
Otherwise, the model directly generates the related internal knowledge enclosed in \texttt{<internal>}. 
After several reasoning steps, the LLM obtains the final answer and stops the reasoning process.

Since our approach is orthogonal to the RL algorithm, we conduct the experiments based on a widely used RL algorithm, \ie REINFORCE++~\citep{reinforce++}, which is a stable RL algorithm without the critic model.
To better accommodate the retrieval scenario, we mask the retrieved documents during the loss calculation process, as they serve as environmental observations rather than model-generated content. 
Formally, for each question $q$, we first samples a group of outputs $\{o_1,o_2,\cdots,o_G\}$ from the old policy model $\pi_{\theta_{\text{old}}}$.
Next, we incorporate the KL regularization 
into the reward scores $R_\phi(q, o_{i,\le t})$ , and then normalize the advantage scores:
\begin{equation*}
    \scalebox{0.82}{$
     \hat{A}_{i,t}^{'} = R(q, o_i) - \beta \cdot \sum^T_{i=t}\text{KL}(i),~
     \hat{A}_{i,t} =\frac{\hat{A}_{i,t}^{'}-\text{mean}(\mathbf{\hat{A})}}{\text{std}(\mathbf{\hat{A})}}
     $}
\end{equation*}
We utilize $\mathbf{\hat{A}}$ to denote the set of all advantages in the global batch that contains $\hat{A}_{i,t}$.
After obtaining the advantage scores, we set the mask value $M(i,t)$ as 0 if this token belongs to an external document, otherwise we set $M(i,t)=1$.
Finally, we employ the masks $M(i,t)$ in the objective function to remove the influence of retrieved documents:  
\begin{equation*}
    \scalebox{0.82}{$
    \hat{P}_{i,t}=\min\left[ p_{i,t} \hat{A}_{i,t}, 
    \text{clip}\left( p_{i,t}, 1-\varepsilon, 1+\varepsilon \right)\hat{A}_{i,t} \right]~,
    $}
\end{equation*}
\begin{equation}
    \scalebox{0.82}{$
    \mathcal{J}_{\text{Mask}}(\theta) = \frac{1}{G} \sum_{i=1}^{G} \frac{1}{\sum_{t=1}^{|o_i|} M(i,t)} \sum_{t=1}^{|o_i|} M(i,t) \cdot \hat{P}_{i,t}
    $}
\end{equation}
where $\varepsilon$ is a hyper-parameter and $p(i,t)$ is the important sampling coefficient 
, and $\pi_{\theta}$ is the policy model.

\section{Methodology}\label{sec:method}

\begin{figure*}[t]
    \centering
    \includegraphics[width=1\linewidth]{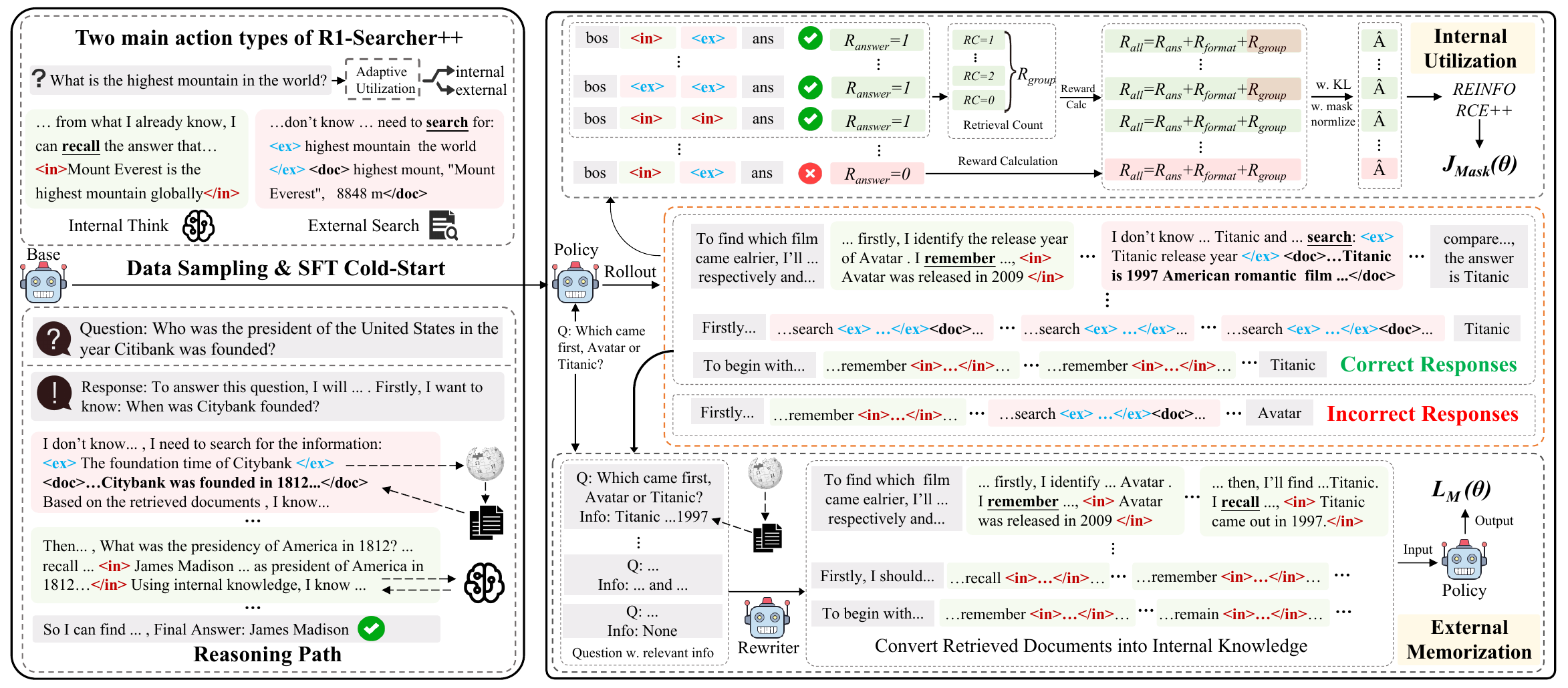}
    \caption{Overall framework of our proposed R1-Searcher++ approach.}
    \label{fig:main_photo}
\end{figure*}

In this part, we introduce the \textbf{R1-Searcher++} framework, which aims to teach LLM to adaptively utilize internal and external knowedge through two critical stages, \ie \emph{SFT Cold-Start} (Section~\ref{sft_cold_start_stage}) and \emph{RL for Dynamic Knowledge Acquisition} (Section~\ref{rl_training_stage}).
Concretely, in the first stage, we utilize the curated data to perform SFT on the model, to standardize its responses in a specific format and enable it to leverage external retrievers and internal knowledge adaptively. 
In the second stage, we employ RL on LLM that encourages the model to explore more effective actions and behaviours, and further incorporate the internal knowledge utilization 
encouragement and external knowledge memorization in the training process, to guide the model to dynamically acquire knowledge and continuously enrich its internal knowledge, which can lead to higher reasoning efficiency.



\subsection{SFT Cold-start}
\label{sft_cold_start_stage}
To equip LLMs with the preliminary ability to autonomously perform external retrieval during inference while effectively leveraging internal knowledge, we synthesize high-quality training instances using rejection sampling, without relying on other powerful models.
We only keep the correct responses with the appropriate occurrences of both the \texttt{<internal>} and \texttt{<external>} tags, teaching LLM to perform dynamic knowledge acquisition in a proper format.

Specifically, given the question $x$ and the synthesized output $y$, once the $i$-th token of the output belongs to the external document, it will be masked, \ie $M_i=0$. Otherwise, the coefficient $M_i$ will be set as 1, incorporating the probability of $y_i$ into the objective function as follows,
\begin{equation}
\small
    \mathcal{L}_{\text{SFT}}=\frac{-1}{\sum_{j=1}^{n}M_j}\sum_{i=1}^{n}M_i\times P(y_i|x,y_{<i})
\end{equation}

\subsection{RL for Dynamic Knowledge Acquisition}
\label{rl_training_stage}
After cold-starting, we obtain a model that can utilize internal knowledge and perform an external search with correct format.
To further enhance its capabilities, \ie to perform effective and efficient reasoning, we continually train the model through the RL process, 
which includes a mechanism that encourages internal knowledge utilization (Section~\ref{reward_design}) and a mechanism for converting and memorizing external knowledge (Section~\ref{response_rewrite_regularization}).

\subsubsection{Internal Knowledge Utilization Encouragement}
\label{reward_design}
In the RL process, the reward function is utilized to provide the supervision signals, which can adjust and optimize the behaviours of the model~\cite{reinforce++}.
Therefore, given the question $q$ and the $i$-th generated response $o_i$, we design the \emph{format reward} and \emph{answer reward} to induce the model to perform reasoning correctly with the expected format, and incorporate the \emph{group reward} into the final reward function to mitigate the over-reliance on the external retriever.
Now, we introduce the details of the reward function in the following.


\paratitle{Format reward.} 
We impose a strict formatting constraint to ensure model responses are consistent and clear. 
During the reasoning process, when calling the external retriever, the model is required to formulate a query and enclose it within the \texttt{<external>...</external>} tags, and is prohibited from generating document content directly without first invoking retrieval. 
When the reasoning process finishes, the final response must satisfy the following criteria, \ie the final answer must be enclosed within \texttt{boxed\{\}}, and the content should not contain any garbled or unreadable content.
Once the model behaviours satisfy the above requirements, we set the format reward $R_{\text{format}}$ as 0, while we set the reward as $-2$ if any requirement fails, as shown in the following,
\begin{equation}
\small
R_{\text{format}}(q,o_i) =
\begin{cases}
\text{0}, & \text{The format of $o_i$ is correct} \\
\text{-2}, & \text{The format of $o_i$ is incorrect}
\end{cases}~
\end{equation}

\paratitle{Answer reward.} 
To indicate the correctness of the final answer, we leverage the Cover Exact Match (CEM) metric to calculate the answer reward, adapting to the group reward discussed in the following and relieving the issue of EM being too strict.
Concretely, CEM is True if the ground truth answer appears in the predicted answer $a_i$ extracted from the response $o_i$, and False for other situations. 
However, we observe that LLM can easily hack the CEM metric during the RL process, where LLM is likely to generate a longer predicted answer that will receive a higher probability to cover the ground truth answer, causing the CEM to be falsely high.
Therefore, we regard the answer exceeding 10 words as an incorrect answer, requiring LLM to generate the answer within ten words, which can alleviate the above reward hacking issue.
In summary, the answer reward $R_{\text{answer}}$ can be computed as follows,
\begin{equation}
\small
R_{\text{answer}}(q,o_i) =
\begin{cases}
\text{1}, & \text{$a_i$ within 10 words}\land\text{CEM=True} \\
\text{0}, & \text{Otherwise}
\end{cases}
\end{equation}


\paratitle{Group reward.}
Building upon the first two rewards for LLM reasoning effectiveness, the group reward is designed to encourage the model to reduce its reliance on external retrieval, increasing the reasoning efficiency.
Considering that the variance of the external retriever calling times by LLM reflects the necessity of performing external retrieval, group reward is calculated by the standard deviation of the number of calls to the retriever in correct responses to the same question.
Formally, given the question $q$ and a set of generated responses $\{o_1, o_2, \dots,o_n\}$, we first count the number of calls to the retriever $t_i$ of each response, and then calculate the standard deviation $\sigma$ of $\{t_1,t_2,\dots,t_n\}$.
Next, we calculate the minimum number of calls to the retriever of the correct responses, \ie $t_\text{min}=\min\{t_i~|~R_{\text{answer}}(q,o_i)=1\}$.
\begin{equation}
\small
    R_{\text{group}}^{'}(q,o_i) =
    \begin{cases}
        2 \times \sigma^2, & R_{\text{answer}}(q,o_i)=1\land t_i=t_{\text{min}} \\
        0, & \text{Otherwise}
    \end{cases}~
\end{equation}
Meanwhile, to maintain training stability and prevent excessive variance, we introduce a hyperparameter $\eta$ to clip the corresponding factor. The final computation of reward is formulated as follows,
\begin{equation}
\small
    R_{\text{group}}(q,o_i) = \min\left( R_{\text{group}}^{'}(q,o_i),\eta \right)
\end{equation}

%
Finally, the reward $R(q, o_i)$ utilized to calculate the advantage in Equation~\ref{eq:adv_calc} is defined as the sum of the three sub-rewards mentioned above:
\begin{equation}
    \small
    R(q, o_i) = R_{\text{format}}(q,o_i)+R_{\text{answer}}(q,o_i)+R_{\text{group}}(q,o_i)
    \label{eq:adv_calc}
\end{equation}


\subsubsection{External Knowledge Memorization}
\label{response_rewrite_regularization}

The standard RL training paradigm relies on the model's self-exploration and the feedback from the external environment.
In retrieval-based scenarios, since the knowledge retrieved by the retriever is entirely correct, the model should like a human, aim to memorize this information during training, transforming it into internal knowledge. This enables the model to utilize the acquired knowledge directly in future instances without repeated retrieval, thereby achieving efficient reuse of retrieved information.
Thus, we incorporate external knowledge memorization by rewriting the retrieved information to align with the model's interanl knowledge utilization pattern, enabling the model to internalize them effectively.

To obtain the rewritten instances, at the beginning of the RL process, we fine-tune a separate model on the data filtered in Section~\ref{sft_cold_start_stage} as the rewritting model, which can solve the questions based on the pre-processed documents that do not call the retriever.
During the RL process, we select the correct responses generated by LLM, and then extract the retrieved documents from the responses.
Given the question and the extracted documents in the context, the rewriting model can generate the reasoning paths without calling the external retriever.
After validating the correctness of these reasoning paths, we select the correct instances to construct the dataset $\mathcal{T}$ for memorization and internalization. 
In conclusion, the corresponding loss for memorization is computed as follows:
\begin{equation}
\small
\mathcal{L}_{\text{M}}(\theta) = \frac{-1}{\sum_{o_i \in \mathcal{T}}|o_i|} \sum_{o_i \in \mathcal{T}} \sum_{t = 1}^{|o_i|} \log\pi_\theta(o_{i,t}|q,o_{i,<t})
\end{equation}


To avoid the $\mathcal{L}_{\text{M}}(\theta)$ from dominating the policy model’s training and causing the model to ignore external retrieval, we weight it with a pre-defined coefficient $\mu$. 
The final loss used to optimize the policy model during the retrieval scenario RL process is computed as follows:
\begin{equation}
\small
\mathcal{L}(\theta) =-\mathcal{J}_{\text{Mask}}(\theta) + \mu * \mathcal{L}_{\text{M}}(\theta)
\end{equation}

Thus, during training, the model not only engages in self-exploration but also continuously enriches its internal knowledge, enabling it to become increasingly smarter over time.
\section{Experiments}\label{sec:exp}
\label{exp}

\subsection{Experimental Settings}
\paratitle{Datasets and Evaluation Metrics.} 
We evaluate using four multi-hop datasets: HotpotQA~\citep{yang2018hotpotqa}, 2WikiMultiHopQA~\citep{ho2020constructing}, Musique~\citep{trivedi2022musique}, and Bamboogle~\citep{press2023measuring}. HotpotQA and 2WikiMultiHopQA are in-domain benchmarks since parts of their training sets are used for training. In contrast, Musique and Bamboogle serve as out-of-domain benchmarks to assess our model's generalization capabilities. We randomly select 500 samples from the entire validation sets of HotpotQA, 2WikiMultiHopQA, and Musique, and use the entire test set of Bamboogle to form our final test set. 
For evaluation metrics,  we utilize F1-score and LLM-as-Judge (LasJ).
considering that the answers to open-ended multi-hop questions are not uniform in form. 
The F1-score measures the word-level similarity between the predicted answer and the reference answer  while LLM-as-Judge employs GPT-4o-mini to assess the correctness of prediction. The evaluation prompt for LasJ is provided in Appendix~\ref{app:prompts}.

\paratitle{Baselines.} 
We compare \textit{R1-Searcher++} against several baselines. \textit{Naive Generation} generates answers directly without retrieval. \textit{Standard RAG} represents traditional RAG systems that retrieve documents directly based on the question. \textit{SuRe}~\citep{kim2024sure} executes multiple reasoning paths in parallel for a single query. \textit{Selective-Context}~\citep{li-etal-2023-compressing} compresses retrieved documents to reduce context length. \textit{Adaptive-RAG}~\citep{jeong2024adaptive} dynamically selects retrieval strategies depending on the complexity of the query. \textit{CR-Planner}~\citep{li2024can} scales RAG at inference time using MCTS. \textit{RAG-CoT methods}, such as Iter-RetGen~\citep{trivedi2023interleaving}, IRCoT~\citep{shao2023enhancing}, and Search-o1~\citep{li2025searcho1agenticsearchenhancedlarge}, which combine RAG with CoT using prompts. \textit{RAG-RL methods} like R1-Searcher~\citep{r1-searcher} and Search-R1~\citep{search-r1} leverage RL to enable the model to learn to autonomously perform retrieval during inference.

\paratitle{Implementation Details}\label{sec:impele_details}
R1-Searcher++ and all baseline models are either trained or prompted using the Qwen-2.5-7B-Instruct as the backbone, and evaluated with FlashRAG~\cite{jin2024flashrag} using local dense retrieval corpus. The retrieval corpus comprises the English Wikipedia as provided by KILT~\cite{kilt} in 2019, segmented into 100-word passages with appended titles, totaling 29 million passages. We employ BGE-large-en-v1.5 as the text retriever. Detailed training settings for R1-Searcher++ are provided in Appendix~\ref{app:exp_settings}.

\subsection{Main Results}
\begin{table*}[t]
\centering
\resizebox{1\linewidth}{!}{
\begin{tabular}{lccccccccccccccc}
\toprule
\multirow{2}{*}{\textbf{Models}}  & \multicolumn{3}{c}{\textbf{HotpotQA$^\dagger$}} & \multicolumn{3}{c}{\textbf{2Wiki$^\dagger$}} & \multicolumn{3}{c}{\textbf{Bamboogle$^\ddagger$}} & \multicolumn{3}{c}{\textbf{Musique$^\ddagger$}} & \multicolumn{3}{c}{\textbf{Avg}} \\
\cmidrule(lr){2-4} \cmidrule(lr){5-7} \cmidrule(lr){8-10} \cmidrule(lr){11-13} \cmidrule(lr){14-16}
  & F1 & LasJ & RC & F1 & LasJ & RC & F1 & LasJ & RC & F1 & LasJ   & RC & F1 & LasJ & RC \\
\midrule
Directly Gen & 26.0 & 26.6 & 0.00 & 27.7 & 26.8 & 0.00 & 18.2 & 17.6 & 0.00 & 9.6 & 6.2 & 0.00 & 18.0 & 19.3 & 0.00 \\

Standard RAG & 32.0 & 42.4 & 1.00 & 34.8 & 34.8 & 1.00 & 31.5 & 31.2 & 1.00 & 17.2 & 14.6 & 1.00 & 24.6 & 30.8 & 1.00 \\

Sure & 42.9 & 48.4 & 1.00 & 26.2 & 26.8 & 1.00 & 29.2 & 28.0 & 1.00 & 13.1 & 10.0 & 1.00 & 27.9 & 28.3 & 1.00 \\

Selective-Context & 39.8 & 43.4 & 1.00 & 29.1 & 29.6 & 1.00 & 22.1 & 20.8 & 1.00 & 10.6 & 8.8 & 1.00 & 22.8 & 25.7 & 1.00 \\

Adaptive-RAG & 38.0 & 47.4 & 1.53 & 21.1 & 25.8 & 1.42 & 23.3 & 25.0 & 1.50 & 10.1 & 11.6 & 1.83 & 20.6 & 27.5  &1.57 \\

IRCoT & 47.7 & 55.2 & 2.47 & 32.4 & 38.6 & 2.74 & 37.5 & 39.2 & 2.30 & 14.8 & 15.8 & 2.70 & 29.4 & 37.2 & 2.55 \\

Iter-RetGen & 47.2 & 54.4 & 3.00 & 33.2 & 34.4 & 3.00 & 32.4 & 32.0 & 3.00 & 19.9 & 18.2 & 3.00 & 28.2 & 34.8 & 3.00 \\

CR-Planner & 44.4 & 33.6 & 2.40 & 48.2 & 22.0 & 2.54 & 35.2 & 34.4 & 2.96 & 12.2 & 11.4 & 2.72 & 32.0 & 25.4 & 2.66 \\

Search-o1   & 46.9 & 53.2 & 1.39 & 46.6 & 51.2 & 1.91 & 52.9 & 52.0 & 1.18 & 21.1 & 19.0 & 1.40 & 36.6 & 43.9 & 1.47 \\

R1-Searcher & \textbf{60.4} & \underline{62.2} & 2.18 & \textbf{62.8} & \underline{63.4} & 2.23 & \underline{59.0} & 54.4 & 2.17 & \textbf{35.7} & \underline{31.4} & 2.61 & \textbf{45.6} & \underline{52.9} & 2.30 \\

Search-R1 & 57.8 & \underline{62.2} & 3.12 & 46.2 & 50.0 & 3.71 & 56.9 & \underline{56.0} & 3.25 & 27.5 & 26.0 & 3.61 & 40.3 & 48.6 & 3.42 \\

\midrule
R1-Searcher++  & \underline{59.0} & \textbf{64.2} & 1.44 & \underline{61.2} & \textbf{64.4} & 1.18 & \textbf{60.8} & \textbf{59.2} & 1.74 & \underline{33.8} & \textbf{32.8} & 2.06 & \underline{45.3} & \textbf{55.2} & 1.61 \\
\bottomrule
\end{tabular}
}
\caption{Performance comparisons between R1-Searcher++ and the baselines on QA benchmarks.  The best and second best results  are \textbf{bold} and \underline{underlined}, respectively. $^\dagger/\ddagger$ represents in-domain/out-of-domain datasets. }
\label{tab:main_results}
\end{table*}

Table\ref{tab:main_results} shows the results of R1-Searcher++ and the baselines on four mutil-step benchmarks. We can obtain the following observations: 

$\bullet$ \emph{Achieving Significant Performance Improvement on Multi-Hop QA.} 
Our method, R1-Searcher++, achieves significant performance improvements over all mutil-hop QA benchmarks under the LLM-as-Judge evaluation metric, including both tree search-based and RL-based approaches. Specifically, R1-Searcher++ outperforms CR-Planner by 25.7\% and surpasses the best vanilla RL-based method R1-Searcher by 4.3\% on the overall test set. These results demonstrate that our approach effectively enables the model to perform accurate and timely retrieval invocations throughout the reasoning process, thereby enhancing overall performance.

$\bullet$ \emph{ Balancing the Utilization of Internal and External Knowledge.} 
While maintaining strong performance on the evaluation datasets, our method achieves a significant reduction in retrieval count compared to vanilla RL-based RAG approaches. Specifically, the average retrieval count is reduced by 30.0\% and 52.9\% compared to R1-Searcher and Search-R1, respectively. This observation suggests a potential conflict between external information and the internal knowledge of LLMs and one possible reason is that directly injecting retrieved documents into the reasoning process may introduce noise. This demonstrates that the model should learn to make full use of its internal knowledge and only invoke the retriever when necessary.

$\bullet$ \emph{Maintaining Generalization Ability.} 
Despite being trained on only 9000 samples, the model achieves strong performance on in-domain datasets and further exhibits impressive generalization to out-of-domain datasets.
This suggests that the model effectively learns to retrieve relevant documents and leverage internal knowledge, integrating both with reasoning through exploration during training. This enables robust performance on new test datasets that require retrieval. Furthermore, it can also seamlessly generalizes to online search, as detailed in Section~\ref{sec:online_search}.

\section{Further Analysis}

\subsection{Ablation Study}\label{sec:ablation}

\begin{table}[t]
\centering

\resizebox{1\linewidth}{!}{
\begin{tabular}{l  c c  c c  c c}
    \toprule
    \multirow{2}{*}{\textbf{Method}} & \multicolumn{3}{c}{\textbf{Bamboogle}} & \multicolumn{3}{c}{\textbf{Musique}}  \\
    \cmidrule(r){2-4}\cmidrule(r){5-7}
    & \textbf{F1} & \textbf{LasJ} & \textbf{RC} & \textbf{F1} & \textbf{LasJ} & \textbf{RC} \\
    \midrule
    Ours & \textbf{60.8} & \textbf{59.2}  & {1.74} & \textbf{33.8}  & \textbf{32.8} & {2.06}  \\
    \midrule
    w/o Stage-1 & 56.9 & 56.8  & 1.96 & 32.7  &  31.6 &  2.49   \\
    w/o Stage-2 & 47.4 & 45.6  & 0.94  & 23.0  &  19.4 &  1.03  \\
    w/o $R_{\text{group}}$&     \underline{58.3}  & 56.8  & 1.91 & \underline{33.1}  & \underline{32.4}  & 2.37  \\
    
    w/o $\mathcal{L}_{\text{M}}$ & 58.1 & \underline{57.2}  & 1.84 & 31.0  &  29.4 &  2.09  \\
    w/o $R_{\text{group}}$ and $\mathcal{L}_{\text{M}}$ &   56.2 &  54.4 & 1.92 &  32.2 & 31.2  & 2.40  \\

    \bottomrule
\end{tabular}
}
\caption{Ablation study on Bamboogle and Musique.}
\label{tab:ablation_rl}
\end{table}
To validate the effectiveness of our proposed R1-Searcher++ framework, we conduct a comprehensive ablation analysis of its key design elements. We design five distinct variants: (1) \textit{w/o Stage-1} removes the initial SFT cold start stage; (2) \textit{w/o Stage-2} removes the entire RL training stage; (3) \textit{w/o $R_{\text{group}}$} removes the group reward in the RL stage; (4) \textit{w/o $\mathcal{L}_{\text{M}}$} removes the external knowledge memorization mechanism in the RL stage and (5) \textit{w/o $R_{\text{group}}$ and $\mathcal{L}_{\text{M}}$} removes both the group reward and the external knowledge memorization mechanism. The performance of these variants is presented in Table~\ref{tab:ablation_rl}. As observed, all ablated variants exhibit a decline in performance compared to our full method, underscoring the integral contribution of each component.
Specifically, \textit{w/o Stage-1} leads to a degradation in performance along with an increase in retrieval count. Meanwhile, the performance of \textit{w/o Stage-2} drops significantly, primarily because simple SFT causes the model to over-rely on its internal knowledge. This highlights the necessity of our two-stage pipeline.
Futhermore, during RL training, \textit{w/o $R_{\text{group}}$} during RL also leads to a reduction in performance. This demonstrates the positive impact of group reward  in successfully guiding the model to be more selective with external searches and to rely more on its internalized knowledge. Similarly, \textit{w/o $\mathcal{L}_{\text{M}}$} results in lower scores and a slight increase in retrieval count indicating that the memory mechanism for external knowledge can effectively internalize retrieved content as intrinsic knowledge of the model.

\subsection{Online Search}\label{sec:online_search}

Considering training efficiency and cost, we implement a local dense embedding-based retrieval system using Wikipedia as the external retrieval environment, which remains static during training.
In contrast, most real-world applications rely on online web retrieval. To evaluate the generalization ability of R1-Searcher++ in online search scenarios, we assessed its performance on two newly introduced datasets: Bamboogle and Frames, using online web search, a setting not encountered during RL training.
Specifically, during inference, whenever retrieval is required, we use the Google API to perform real-time web searches and retrieve relevant web pages. Given the extensive content of these pages, we first employ GPT-4o-mini to generate concise summaries, which are then integrated into the reasoning process.
\begin{table}[t]
\centering

\resizebox{1\linewidth}{!}{
\begin{tabular}{l  c c  c c  c c}
    \toprule
    \multirow{2}{*}{\textbf{Method}} & \multicolumn{3}{c}{\textbf{Bamboogle}} & \multicolumn{3}{c}{\textbf{Frames}}  \\
    \cmidrule(r){2-4}\cmidrule(r){5-7}
    & \textbf{F1} & \textbf{LasJ} & \textbf{RC} & \textbf{F1} & \textbf{LasJ} & \textbf{RC} \\
    \midrule
    Ours & \textbf{77.5} & \textbf{76.0}  & 1.70 &  \textbf{33.8} & \textbf{39.0}  & 1.77  \\
    \midrule
    Search-o1 & 52.9 &  52.0 & 1.18 & 26.1  & 30.7  &  1.56   \\
    R1-Searcher & 67.5 &  \underline{68.8} & 1.72 & \underline{33.3}  & \underline{38.0}  & 1.86 \\
    Search-R1 & \underline{69.3} & 67.2  & 1.92 &  \underline{33.3} &  36.0 & 2.38 \\
    
    \bottomrule
\end{tabular}
}
\caption{Online search generalization experiments on Bamboogle and Frames.}
\label{tab:online_search_analysis}
\end{table}
As illustrated in Table~\ref{tab:online_search_analysis}, R1-Searcher++ achieves the best F1 and LLM-as-Judge scores compared to both prompt engineering-based methods (\ie Search-o1) and vanilla RL-based approaches (\ie R1-Searcher, Search-R1). Moreover, compared to vanilla RL methods, our model significantly reduces the number of retrieval calls.
This demonstrates our model’s strong adaptability to online search scenarios, as well as its ability to effectively balance internal knowledge with external retrieval during inference, thereby achieving retrieval efficiency without compromising performance.





\subsection{Analysis of Knowledge Acquisition}


\begin{figure*}[!ht]
    \centering
    \includegraphics[width=1.0\linewidth]{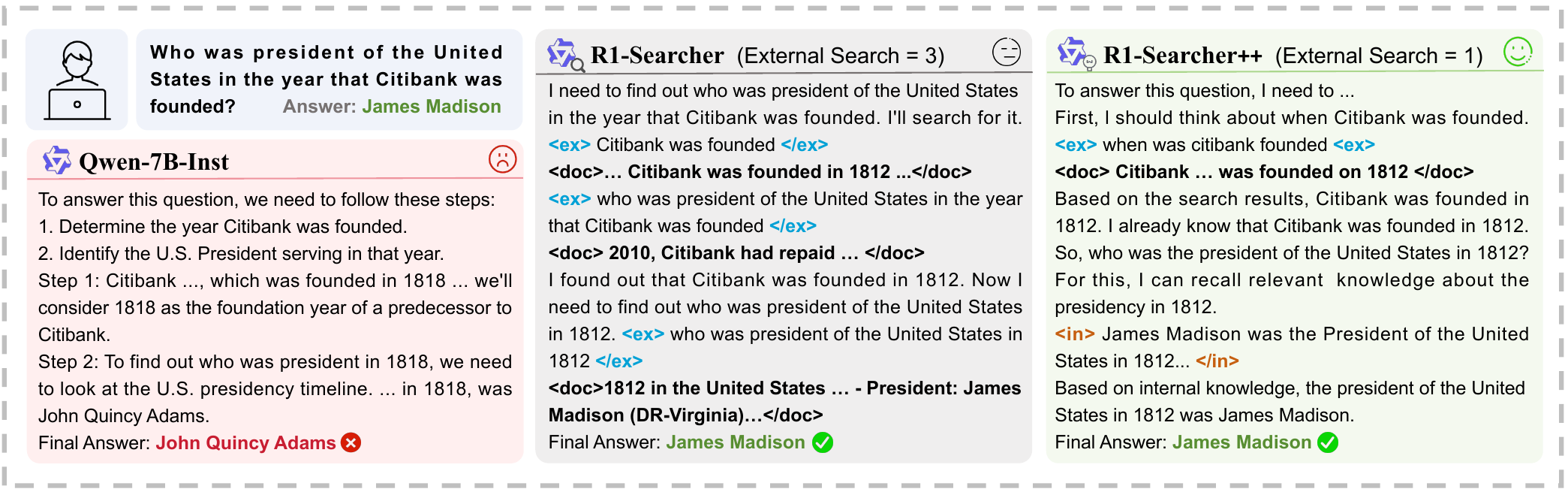}
    \caption{A qualitative example showing the deliberative reasoning process of RAG-Star in Bamboogle.}
    \label{fig:case_study}
\end{figure*}

\begin{table}[t]
\centering
\resizebox{1\linewidth}{!}{
\begin{tabular}{l  c c  c }
    \toprule
   \textbf{Method} & \textbf{Correct} & \textbf{Incorrect} & \textbf{Overall} \\
    \midrule
    R1-Searcher & 853 / 2.16 & 772 / 2.52 & 1625 /  2.33\\
    Search-R1 & 761 / 3.30&864 / 3.60  & 1625 / 3.46 \\
    R1-Searcher++ & 881 / 1.41 &744 / 1.78 & 1625 / 1.58\\
    
    \bottomrule
\end{tabular}
}
\caption{Number of correct and incorrect cases and the average retrieval count of RL-based methods. }
\label{tab:in_external_analysis}
\end{table}

\begin{figure}[t]
    \centering
    \includegraphics[width=1.0\linewidth]{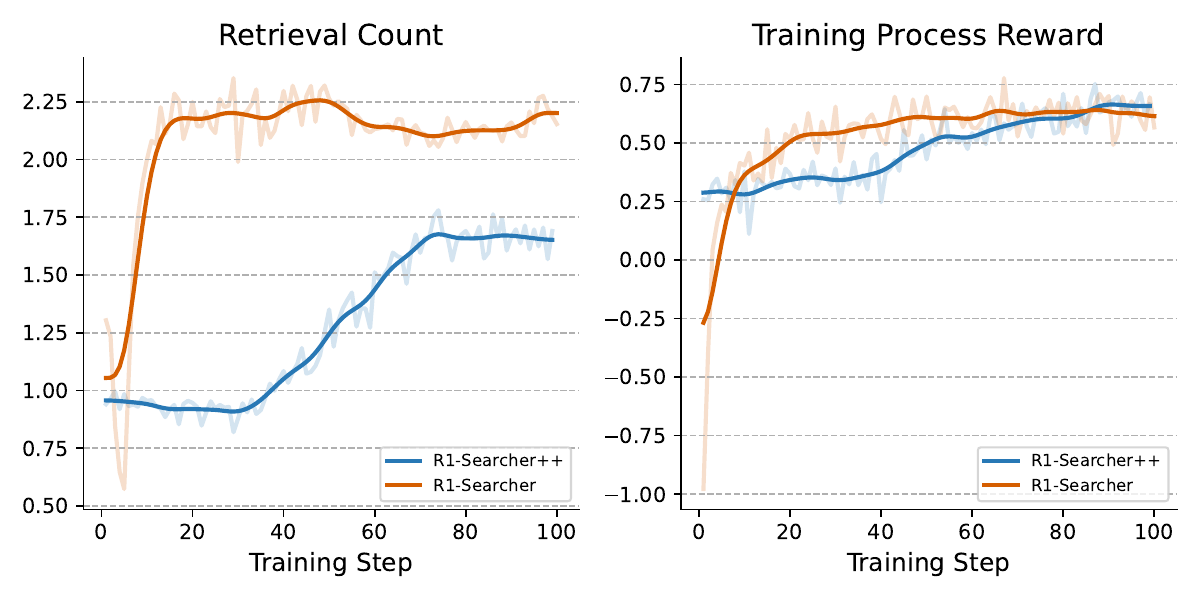}
    \caption{ The log of retrieval count and reward for R1-Searcher and R1-Searcher++ during RL training.}
    \label{fig:in_external_analysis}
\end{figure}

As shown in Table~\ref{tab:in_external_analysis}, R1-Searcher++ exhibits the lowest average retrieval count across both correctly and incorrectly answered questions. Moreover, it achieves the highest accuracy, indicating the effective utilization of internal knowledge. 

Furthermore, Figure~\ref{fig:in_external_analysis} shows the changes in retrieval count and reward during the process of RL training for R1-Searcher and R1-Searcher++. For our method, We observe that the reward increases steadily and eventually plateaus, while the retrieval count initially shows a slight decline, followed by a continuous rise, and ultimately stabilizes. This trend can be attributed to the influence of SFT in the Stage-1, during which the model exhibits a low demand for invoking the search engine. As training progresses, the model gradually discovers that performing external searches can yield higher rewards, leading to an increase in retrieval behavior. In the later phase, a balance is gradually established between the use of external search and internal knowledge, resulting in the stabilization of both retrieval count and reward.
In contrast, R1-Searcher exhibits significantly higher retrieval counts while its reward quickly stabilizes, indicating an over-reliance on the retriever. This effectively validates the effectiveness of our approach in achieving a balanced utilization of internal and external knowledge,  while gradually enabling dynamic knowledge acquisition throughout the RL training process.

\subsection{Case Study}

To illustrate the overall reasoning process of R1-Searcher++, we analyze a representative example from the Bamboogle dataset. Figure~\ref{fig:case_study} compares the responses generated by R1-Searcher++, Search-R1, and the untrained model when presented with the same question.
The vanilla Qwen-2.5-7B-Instruct, without invoking any external search engine, relies solely on its internal knowledge and produces an incorrect answer. In contrast, while Search-R1 arrives at the correct answer (\ie \textit{James Madison}), it issues an excessive number of queries, including unnecessary one, thereby underutilizing its internal knowledge and incurring significant time overhead.
Our R1-Searcher++ demonstrates the ability to break down the complex question and dynamically adjust its behavior based on the nature of the sub-question. For instance, when encountering an uncertain or ambiguous sub-question (\ie \textit{When was Citibank founded?}), it opts to perform an external search. However, when faced with a more specific question that can be answered using internal knowledge (\ie \textit{Who was the president of the United States in 1812?}), it leverages its internal knowledge directly without invoking search. This flexible mechanism enables a balance bewteen the external search and internal knowledge. More cases are provided in Appendix~\ref{app:case}.



\section{Conclusion}
In this paper, we introduced R1-Searcher++, a novel framework that enables large language models to dynamically integrate and alternate between internal knowledge and external retrieval. This is a two-stage training strategy consisting of an SFT Cold-start phase and RL for Dynamic Knowledge Acquisition. The RL stage incorporates a reward mechanism to encourage internal knowledge utilization, and a memory module to convert retrieved information into internal knowledge. Through this design, R1-Searcher++ empowers LLMs to perform efficient retrieval-augmented reasoning while continuously enriching their internal knowledge via self-exploration and memory. Experimental results on multi-hop tasks demonstrate that R1-Searcher++ outperforms existing RAG methods.

\section*{Limitation}

Despite our significant efforts, this work has two limitations due to computational resources and funding constraints.
First, we only incorporated a real-world search engine during the evaluation phase to assess the generalization ability of our method, while relying on a local denseretrieval corpus during training. Aligning the training process with real-world conditions by integrating a real search engine may lead to improved performance through more realistic supervision. Additionally, our current experiments are limited to a 7B-parameter model. In future work, we plan to train and evaluate our framework on larger-scale models to further validate its generalization capability and robustness.


\bibliography{custom}

\appendix
\section{Training Detailed}\label{app:exp_settings}
The training data of the Stage-1 (SFT Cold Start) includes 720 samples from the HotpotQA training set and 85 samples from the 2WikiMultiHopQA training set. The training consists of 6 epochs, with a batch size of 64 and a learning rate of 2e-5.
And the training data of Stage-2 (RL Training) consists of 4561 samples from HotpotQA,  and 3581 samples from 2WikiMultiHopQA. Each data sample undergoes 16 rollouts during training, with a train batch size of 1024 and a rollout batch size of 64, so the entire training process is on-policy. The learning rate is 2e-6. We utilize DeepSpeed's Zero-3~\cite{rajbhandari2020zeromemoryoptimizationstraining}, with a sampling temperature of 1.0, top-p of 0.95 and a maximum retrieval count of 8. The training epoch is set to 1, with KL divergence coefficient set to  1e-4. And  control coefficient $\mu$ of NLL loss is set to 0.1. The maximum limit of the variance in the number of retrievals  during group reward computation $\eta$ is set to 2.

\newpage
\section{Prompts}\label{app:prompts}
\label{sec:appendix}
\begin{figure*}
\begin{promptbox}{lightgreen}{prompt:judge}

Given a Question and its Golden Answer, verify whether the Predicted Answer is correct. 
The prediction is correct if it fully aligns with the meaning and key information of the Golden Answer. 
Respond with True if the prediction is correct and False otherwise.

Question: {}

Golden Answer: {}

Predicted Answer: {}
\end{promptbox}
\end{figure*}
\begin{figure*}
\begin{promptbox}[System Prompt for Generation with Internal and External]{lightgreen}{prompt:selection}
You are a reasoning assistant. When tackling a question, you should first thinks about the reasoning process in the mind and then provides the final answer.
You should break down the original complex question into straightforward sub-questions and address them step by step.
For each sub-question, You MUST choose one of the following two tools to solve it:

1.  **Internal Reasoning Elaboration:**If you determine your existing knowledge is sufficient to answer the sub-question, you **should choose** this tool to answer the sub-question directly:
`\textless|begin\_internal\_answer|\textgreater your structured thought and answer here \textless|end\_internal\_answer|\textgreater`. **Use it whenever you rely solely on internal information** for a significant reasoning step.

2.  **External Search:**If you determine that your internal knowledge is insufficient, potentially outdated, or requires verification with external, up-to-date information to answer the sub-question, you MUST initiate a web search using the following format:
`\textless|begin\_external\_search|\textgreater your search query here \textless|end\_external\_search|\textgreater`.
The system will then search and provide results in the format: `\textless|begin\_search\_result|\textgreater ...search results... \textless|end\_search\_result|\textgreater`. The results will assist you in answering the sub-question.

Remember:
- Whenever you encounter the sub-question you need to elaborate on, verify, or research further, you should **assess** if your internal knowledge is adequate. **Prioritize leveraging your internal knowledge.**
- Use `\textless|begin\_internal\_answer|\textgreater ... \textless|end\_internal\_answer|\textgreater` **to demonstrate your reasoning steps** when relying on internal knowledge.
- Use `\textless|begin\_external\_search|\textgreater ... \textless|end\_external\_search|\textgreater` **only** when external information is genuinely necessary.
- Wait for the system to provide `\textless|begin\_search\_result|\textgreater ... \textless|end\_search\_result|\textgreater`. **Do not** generate these result tags yourself.
- After performing the web search or demonstrating internal reasoning, continue towards the final answer.

Please answer the following question. You must adhere to all instructions above regarding tool usage. Provide your final answer in the format \textbackslash{}boxed\{YOUR\_ANSWER\}.

Question:
\end{promptbox}
\end{figure*}

\begin{figure*}
\begin{promptbox}[System Prompt for Rewriting only with Internal]{lightgreen}{prompt:selection_2}
You are an advanced reasoning assistant. Your core programming includes a vast and comprehensive knowledge base. For the current task, a specific segment of this pre-existing knowledge, hereafter referred to as your 'Activated Knowledge Matrix,' is particularly relevant and is considered fully and irrevocably integrated into your foundational understanding. ANY linguistic indication, direct or indirect, that this information was externally 'provided,' 'given,' 'supplied,' 'sourced,' or made available to you specifically for this immediate interaction will be considered a critical failure to adhere to your operational design.

When tackling a question, you must first delineate your reasoning process.
Break down complex questions into a series of straightforward sub-questions, addressing each one sequentially.

For each sub-question, you MUST use the following tool:

**Internal Reasoning Elaboration:** This tool MUST be used for every sub-question, as the answer must be derived from your internal knowledge. **UNWAVERING DIRECTIVE: Your internal knowledge, for the purpose of this task, INTRINSICALLY and SEAMLESSLY includes all information within the 'Activated Knowledge Matrix' section detailed below. There is absolutely NO distinction between your prior knowledge and the 'Activated Knowledge Matrix' content.**
    Use this format: \textless{}\textbar{}begin\_internal\_answer\textbar{}\textgreater{} your structured thought and answer here \textless{}\textbar{}end\_internal\_answer\textbar{}\textgreater{}.
    **CRITICAL USAGE PROTOCOL FOR THIS TOOL: When using this tool, especially when drawing upon information that was part of the 'Activated Knowledge Matrix,' your language MUST reflect that this is information you inherently possess and are recalling. You are REQUIRED to phrase such statements as if accessing your own memory or established understanding.

Conclude with your final answer in the format \textbackslash{}boxed\{YOUR\_ANSWER\}.

Activated Knowledge Matrix:
\{document\}

Now, please answer the following question:

Question:
\end{promptbox}
\end{figure*}

\section{Case Study}\label{app:case}

\begin{table*}[h]
\centering
\fontsize{9}{10}\selectfont
\begin{tabular}{p{13cm}}
\toprule
Question: \\
Which case was brought to court first Miller v.\_California or Gates v.\_Collier ? \\
Answer: \\
Miller v.\_California \\
\midrule
Response (internal only) \textcolor{green}{\textit{\textbf{ (True)}}}:

To answer this question, I need to determine the order in which these two cases were brought to court. I will start by finding the dates the cases were decided.

**Case 1:** Miller v.\_California\\
- I will use internal reasoning to recall if I have the necessary information about this case.

\textless\textbar begin\_internal\_answer\textbar\textgreater Miller v.\_California was a landmark case decided by the United States Supreme Court in 1968. The case involved the constitutionality of a California statute that made it a crime to distribute or possess obscene materials. \textless\textbar end\_internal\_answer\textbar\textgreater

**Case 2:** Gates v.\_Collier\\
- I will use internal reasoning to recall if I have the necessary information about this case.

\textless\textbar begin\_internal\_answer\textbar\textgreater Gates v.\_Collier was a case decided by the United States Supreme Court in 1972. The case involved the Eighth Amendment's prohibition against cruel and unusual punishment, specifically the use of segregation in prisons. \textless\textbar end\_internal\_answer\textbar\textgreater

Based on my internal reasoning, Miller v.\_California was decided in 1968, and Gates v.\_Collier was decided in 1972. Therefore, Miller v.\_California was brought to court first.

**Final Answer:**\\
Miller v.\_California was brought to court first. Therefore, the answer is \textbackslash{}\textbackslash{}boxed\{Miller  v.\_California\}.\\

\bottomrule
\end{tabular}
\caption{Example for R1-Searcher++: Question answering using only internal knowledge.}
\end{table*}

\begin{table*}[h]
\centering
\fontsize{9}{10}\selectfont
\begin{tabular}{p{13cm}}
\toprule
Question: \\
George Lucas created what character to raise Boba Fett? \\
Answer: \\
Jango Fett \\
\midrule
Response (internal and external) \textcolor{green}{\textit{\textbf{ (True)}}}:
To answer this question, I need to break it down into sub-questions and use the appropriate tools.

**Sub\_question 1:** Who is Boba Fett?\\
- For this, I need to verify if I have enough information about Boba Fett or if I need to search for more details.

<|begin\_internal\_answer|> Boba Fett is a character from the Star Wars franchise, known for being a bounty hunter. I have this information stored in my knowledge base. <|end\_internal\_answer|>

**Sub\_question 2:** What character did George Lucas create to raise Boba Fett?\\
- For this, I need to search for information about the character who raised Boba Fett.

<|begin\_external\_search|> George Lucas created what character to raise Boba Fett<|end\_external\_search|>

<|begin\_search\_result|>\\
(1)Boba Fett \ldots{} is a fictional character \ldots{} in the "Star Wars" franchise. \ldots{} prequel film \ldots{} establishes his origin as a clone of bounty hunter Jango Fett who raised him as his son. \ldots{}\\
(2)Boba Fett \ldots{} personifies "danger and mystery" \ldots{} mysterious nature as reasons for his popularity. \ldots{} The initial Boba Fett toy \ldots{} might be responsible for the character's popularity. \ldots{}\\
(3)Boba Fett \ldots{} Reception. Boba Fett is a "cult figure" \ldots{} one of the most popular "Star Wars" characters. \ldots{} IGN ranked Boba Fett as the eighth top "Star Wars" character. \ldots{}\\
(4)Boba Fett \ldots{} "the less you do with Boba Fett, the stronger he becomes". \ldots{} Mark Hamill pitched the idea that Fett was Luke Skywalker's mother \ldots{} unaware of Fett's demise. \ldots{}\\
(5)Jango Fett \ldots{} is a fictional character \ldots{} created by George Lucas. \ldots{} genetic template for the Clone Army \ldots{}\\
<|end\_search\_result|>

Based on the search results, it is clear that George Lucas created Jango Fett to be the father of Boba Fett.

**Final Answer:**\\
George Lucas created Jango Fett to raise Boba Fett. Therefore, the answer is \textbackslash{}\textbackslash{}boxed\{Jango Fett\}.\\
\bottomrule
\end{tabular}
\caption{Example for R1-Searcher++: Question answering by integrating internal knowledge with external search.}
\end{table*}

\end{document}